\documentclass[journal,comsoc]{IEEEtran}
\usepackage[T1]{fontenc}
\usepackage{amsmath}
\usepackage[cmintegrals]{newtxmath}
\usepackage{bm}
\usepackage{graphicx}
\usepackage{multirow}
\usepackage{floatrow}
\usepackage{comment}
\usepackage{hyperref}
\usepackage{color}

\usepackage{algorithmic}
\usepackage{algorithm}
\usepackage{bm}
\usepackage{mathtools}
\usepackage{booktabs}
\usepackage{threeparttable}
\usepackage{tabularx}
\newcolumntype{Y}{>{\centering\arraybackslash}X}
\begin{document}
\title{Data Augmentation with norm-VAE for Unsupervised Domain Adaptation}
\author{Qian~Wang,~\IEEEmembership{Member,~IEEE,}
        Fanlin~Meng,
        Toby~P.~Breckon,~\IEEEmembership{Member,~IEEE,}
\thanks{Q. Wang is with the Department
of Computer Science, Durham University, United Kingdom, e-mail: qian.wang173@hotmail.com}
\thanks{F. Meng is with Department of Mathematical Sciences, University of Essex, United Kingdom, e-mail: fanlin.meng@essex.ac.uk}
\thanks{TP. Breckon is with the Department of Computer Science and Department of Engineering, Durham University, United Kingdom, e-mail: toby.breckon@durham.ac.uk}
}

\markboth{Journal of xxx,~Vol.~14, No.~8, August~2020}%
{Wang \MakeLowercase{\textit{et al.}}: Data Augmentation with norm-VAE for Unsupervised Domain Adaptation}
\maketitle

\begin{abstract}
We address the Unsupervised Domain Adaptation (UDA) problem in image classification from a new perspective. In contrast to most existing works which either align the data distributions or learn domain-invariant features, we directly learn a unified classifier for both domains within a high-dimensional homogeneous feature space without explicit domain adaptation. To this end, we employ the effective Selective Pseudo-Labelling (SPL) techniques to take advantage of the unlabelled samples in the target domain. Surprisingly, data distribution discrepancy across the source and target domains can be well handled by a computationally simple classifier (e.g., a shallow Multi-Layer Perceptron) trained in the original feature space. Besides, we propose a novel generative model \textit{norm-VAE} to generate synthetic features for the target domain as a data augmentation strategy to enhance classifier training. Experimental results on several benchmark datasets demonstrate the pseudo-labelling strategy itself can lead to comparable performance to many state-of-the-art methods whilst the use of \textit{norm-VAE} for feature augmentation can further improve the performance in most cases. As a result, our proposed methods (i.e. \textit{naive-SPL} and \textit{norm-VAE-SPL}) can achieve new state-of-the-art performance with the average accuracy of 93.4\% and 90.4\% on Office-Caltech and ImageCLEF-DA datasets, and comparable performance on Digits, Office31 and Office-Home datasets with the average accuracy of 97.2\%, 87.6\% and 67.9\% respectively.
\end{abstract}


\IEEEpeerreviewmaketitle

\section{Introduction}\label{sec:introduction}
\IEEEPARstart{I}{n} the last decade, impressive progress has been made in supervised image classification with the advancement of deep learning \cite{goodfellow2016deep} and particularly deep Convolutional Neural Networks (CNN) trained on large scale image dataset such as ImageNet \cite{deng2009imagenet}. One key to the success of deep neural networks in image classification is the access of sufficient annotated images which are usually unavailable in many real-world applications. To address the issues of training data scarcity in practice, a variety of techniques (e.g., semi-supervised learning \cite{zhu2009introduction}, zero-shot learning \cite{wang2017zero,wang2020multi,keshari2020generalized,schonfeld2019generalized}, domain adaptation \cite{wang2018deep,zhang2019domain,wang2020unsupervised,liang2020we}) can be employed based on the availability of varied training resources. Among these, Unsupervised Domain Adaptation (UDA) assumes the access of labelled data only from the \textit{source domain} where the labelled data are easier to obtain but the data distribution is different from that of the \textit{target domain} in which the task of interest resides. As a result, a classifier trained on the labelled source domain suffers from significant performance drop when directly applied to the target domain. Unsupervised domain adaptation problems are common in real-world applications. For example, recognizing objects in X-ray baggage screening imagery \cite{wang2020generalized} can be a challenging task due to the difficulty of data collection in this domain but regular images are much easier to obtain. In this case, domain adaptation techniques can play a crucial role in making the most of large-scale regular images from the source domain and limited X-ray images from the target domain.

Existing UDA approaches try to align the source and domain data distributions by feature transformation (e.g., projecting features into a subspace) \cite{wang2019unifying,chen2019progressive,wang2020unsupervised} or learning domain-invariant features from images via specially designed deep neural networks \cite{long2015learning,kang2019contrastive}. Subsequently, simple classifiers such as Nearest Neighbour (NN) or Support Vector Machines (SVM) can be employed in the learned domain-invariant feature space. Although impressive performance has been achieved in prior works, we argue that \textit{explicit domain adaptation before learning a classifier is not as necessary as we thought}. To justify this argument, we demonstrate that a unified classifier can be trained for both source and target domain data in a high-dimensional homogeneous feature space despite the existence of domain shift due to the \textit{blessing of dimensionality} \cite{gorban2020high}. From this perspective, the key challenge of UDA problems is the lack of labelled data in the target domain for supervised learning.

\begin{figure}
    \centering
    {\includegraphics[width=\textwidth]{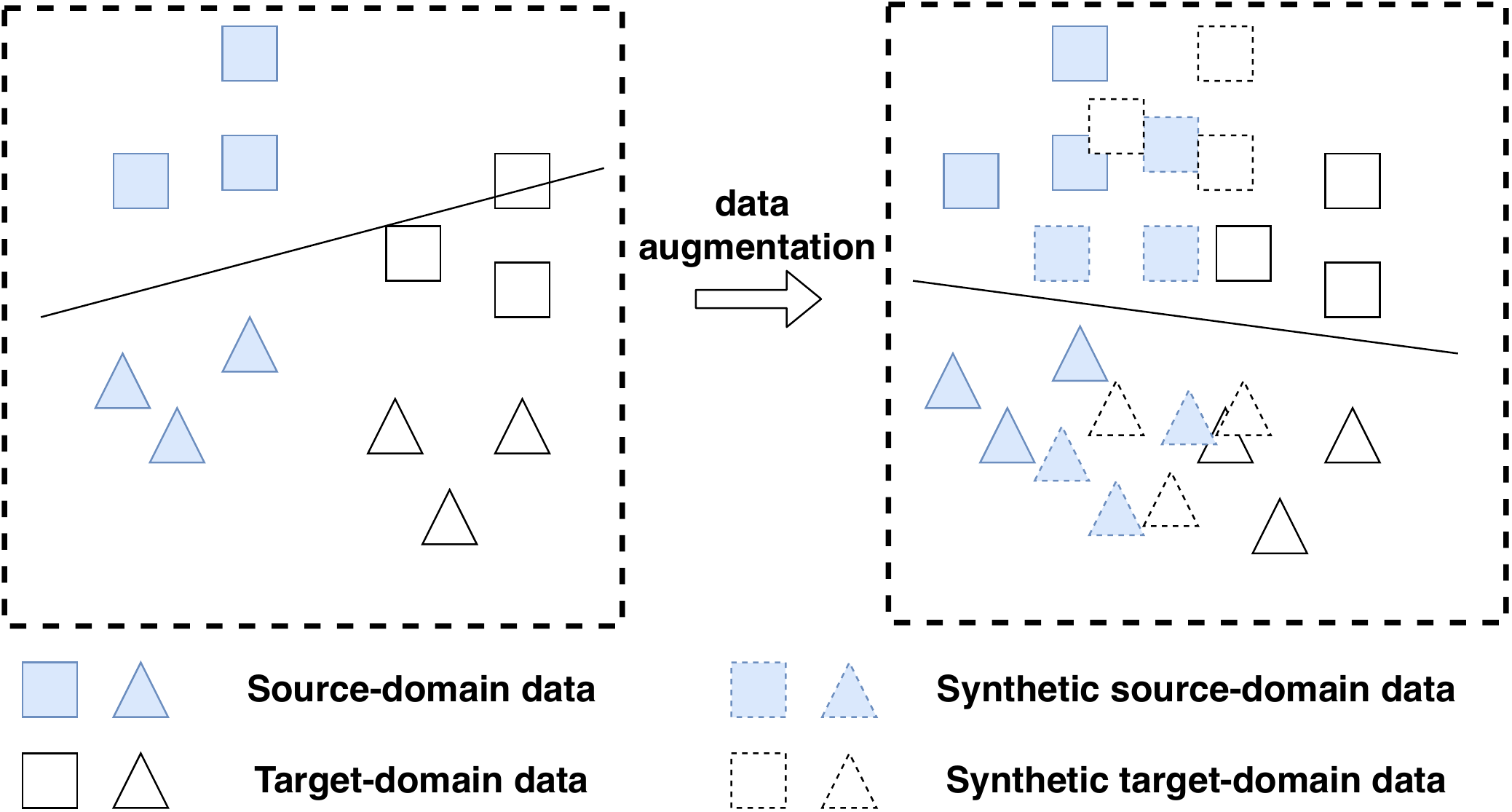}}
    {\caption{An illustration of how data augmentation by synthesizing source and target domain data can benefit unsupervised domain adaptation. Left: classifier trained with labelled source-domain data only; Right: classifier trained with real and synthetic data from both domains.}
        \label{fig:norm-vae-scheme}}
\end{figure}

In this paper, we address unsupervised domain adaptation for image classification from the perspective of target domain data pseudo-labelling and generation. On one hand, we investigate the effectiveness of pseudo labelling techniques without any explicit source and target data distribution alignment. Pseudo labelling techniques have been employed in prior work \cite{wang2019unifying,chen2019progressive,wang2020unsupervised} but its effectiveness has been underestimated. Our experiments demonstrate surprisingly strong classification performance on UDA benchmark datasets with a simple classifier trained on labelled source data and pseudo-labelled target data (e.g., a linear two-layer Multi-Layer Perceptron for image features or a Convolutional Neural Network for digit images). Besides, we propose a novel Variational Autoencoder (i.e. \textit{norm-VAE}) to generate synthetic labelled target samples for training the classifier. The proposed {\it norm-VAE} is characterized by $L2$-normalized parameters (i.e. mean and variance) of latent code distribution as the substitute of the KL-Divergence regularisation in the vanilla VAE. With this data augmentation strategy, the performance of UDA can be enhanced as illustrated in Figure \ref{fig:norm-vae-scheme}.

The contributions of our work can be summarised as follows:
\begin{itemize}
\item[--]{we demonstrate that a specially designed pseudo-labelling strategy can achieve surprisingly strong performance on commonly used benchmark datasets for unsupervised domain adaptation; the performance is even comparable with or better than many more complex state-of-the-art methods on Digits (97.2\%), Office-Caltech (92.8\%) and ImageCLEF-DA (89.4\%).}
\item[--]{we demonstrate the proposed pseudo-labelling strategy is superior to those in \cite{chen2019progressive} and \cite{wang2020unsupervised} within our proposed framework.}
\item[--]{we propose a generative model adapted from VAE to further improve the performance of unsupervised domain adaptation by generating synthetic features for the target domain;  the average accuracy is improved by 0.6\%, 1.7\%, 1\% and 2.2\% on Office-Caltech, Office31, ImageCLEF-DA and Office-Home datasets respectively.}
\item[--]{we present a thorough set of comparative experiments and ablation studies to demonstrate the proposed methods can achieve new state-of-the-art performance on several benchmark datasets (i.e. Digits, Office-Caltech, Office31, ImageCLEF-DA and Office-Home).}
\end{itemize}


\section{Related Work}\label{sec:relatedwork}
In this section, we review existing work related to ours. We first review existing approaches to UDA problems which fall into two main categories: \textit{feature transformation approaches} \cite{long2013transfer, long2014transfer, sun2016return, zhang2017joint, ghifary2017scatter, sun2017correlation, wang2018visual} and \textit{deep feature learning approaches} \cite{ganin2015unsupervised, long2015learning, ganin2016domain, long2016unsupervised, long2017deep, chen2018joint, pei2018multi, zhang2018collaborative, long2018conditional}. Subsequently, we discuss the use of pseudo-labelling and data augmentation techniques in UDA for image classification. Finally, we discuss how Variational Autoencoders are employed for domain adaptation in existing literature.

\subsection{Unsupervised Domain Adaptation} \label{sec:related_uda}

Feature transformation approaches aim to transform the source domain and/or target domain features such that transformed source and target domain data can be aligned. As such the classifier learned from labelled source data can be directly applied to target data. Usually linear transformations are used by learning the projection matrices with different optimization objectives and a kernel trick can help to explore the non-linear relations between source and target domain data if necessary. The most commonly employed objective for unsupervised domain adaptation is to align data distributions in source and target domains \cite{long2013transfer, long2014transfer}. For this purpose, Maximum Mean Discrepancy (MMD) based distribution matching has been used to reduce differences of the marginal distributions \cite{long2014transfer}, conditional distributions or both \cite{long2013transfer, wang2018visual}. 
Correlation alignment (CORAL) \cite{sun2016return} transforms source domain features to minimize domain shift by aligning the second-order statistics of source and target distributions. 
Manifold Embedded Distribution Alignment (MEDA) \cite{wang2018visual} learns a domain-invariant classifier based on the transformed features where the transformation aims to align both the marginal and conditional distributions with quantitative account for their relative importance. 

In contrast to the above mentioned approaches that learns one feature transformation matrix for either source domain or both domains, 
Joint Geometrical and Statistical Alignment (JGSA) \cite{zhang2017joint} learns two coupled projections that project the source and target domain data into a joint subspace where the geometrical and distribution shifts are reduced simultaneously. 
Apart from the distribution alignment, recent feature transformation based approaches also promote the discriminative properties in the transformed features. Scatter Component Analysis (SCA) \cite{ghifary2017scatter} aims to learn a feature transformation such that the transformed data from different domains have similar scattering and the labelled data are well separated. A Linear Discriminant Analysis (LDA) framework was proposed in \cite{lu2018embarrassingly} by learning class-specific projections. Similarly, Li et al. \cite{li2018domain} proposed an approach to feature transformation towards Domain Invariant and Class Discriminative (DICD) features.

Deep feature learning approaches to domain adaptation were inspired by the success of deep Convolutional Neural Networks (CNN) in visual recognition \cite{lecun2015deep}. Attempts have been made to take advantage of the powerful representation learning capability of CNN combined with a variety of feature learning objectives. Most deep feature learning approaches aim to learn domain-invariant features from raw image data in source and target domains in an end-to-end framework. Specifically, the objectives of feature transformation approaches have been incorporated in the deep learning models. To learn the domain-invariant features through a deep CNN, the gradient reversal layer was proposed in \cite{ganin2015unsupervised} and used in other deep feature learning approaches \cite{ganin2016domain, pei2018multi, zhang2018collaborative} as well. The gradient reversal layer connects the feature extraction layers and the domain classifier layers. During backpropagation, the gradients of this layer multiplies a certain negative constant to ensure the feature distributions over two domains are made similar (as indistinguishable as possible for the domain classifier). Deep Adaptation Networks (DAN) \cite{long2015learning} and Residual Transfer Network (RTN) \cite{long2016unsupervised} aim to learn transferable features from two domains by matching the domain distributions of multiple hidden layer features based on MMD. Deep CORAL \cite{sun2016deep} integrates the idea of CORAL \cite{sun2016return} into a deep CNN framework to learn features with favoured properties (i.e. aligned correlations over source and target distributions for multiple layer activations). These approaches only consider the alignment of marginal distributions and cannot ensure the separability of target data. Deep Reconstruction Classification Network (DRCN) \cite{ghifary2016deep} trains a feature learning model using labelled source data and unlabelled target data in the supervised and unsupervised learning manners respectively. More recently, the prevalent Generative Adversarial Network (GAN) loss has been employed in Adversarial Discriminative Domain Adaption (ADDA) \cite{tzeng2017adversarial} with promising results. 

\subsection{UDA With Pseudo-Labelling}\label{sec:pl}
To address the issue of lack of labelled data in the target domain, pseudo-labelling has been used by many existing approaches. Pseudo-labels are assigned to unlabelled samples in the target domain by a classifier.
Hard labelling assigns a pseudo-label $\hat{y}$ to each unlabelled sample without considering the confidence  \cite{long2013transfer,zhang2017joint,wang2018visual}. The pseudo-labelled target samples together with labelled source samples are used to learn an improved classifier. By repeating these two steps, the classifier and accuracy of pseudo-labels can be improved gradually. Hard pseudo-labelling relies heavily on good initialisation otherwise it is likely to be stuck in local optima. 
To address this issue, soft labelling was employed in \cite{pei2018multi}. Instead of assigning a hard label to a sample, soft labelling assigns the probability of belonging to each class to a sample. In the Multi-Adversarial Domain Adaptation (MADA) approach \cite{pei2018multi}, the soft pseudo-label of a target sample is used to determine how much this sample should be attended to different class-specific domain discriminators. 

Selective pseudo-labelling is the other way to alleviate the mis-labelling issue \cite{zhang2018collaborative,wang2019unifying,chen2019progressive}. 
Similar to the soft labelling strategy, selective pseudo-labelling also takes into consideration the confidence in target sample labelling but in a different manner.
Selective pseudo-labelling picks up a subset of target samples and assigns them with pseudo labels with high confidence to avoid potential mis-labelling. The idea is that at the beginning the classifier is weak so that only a small fraction of the target samples can be correctly classified. When the classifier gets stronger after each iteration of learning, more target samples can be correctly classified hence should be pseudo-labelled and participate in the learning process. An easy-to-hard strategy was employed in \cite{chen2019progressive}. Target samples whose similarity scores are higher than a threshold are selected for pseudo-labelling and this threshold is updated after each iteration of learning so that more unlabelled target samples can be selected. A class-wise sample selection strategy was proposed in \cite{wang2019unifying, wang2020unsupervised}. Samples are selected for each class independently so that pseudo-labelled target samples will contribute to the alignment of conditional distribution for each class during learning.
In this paper, we propose a novel pseudo-label selection strategy which is superior to those used in \cite{wang2020unsupervised} within the proposed framework.

\subsection{UDA With Data Augmentation}
Data augmentation has drawn attention in existing works for UDA. For example, Hsu et al. \cite{hsu2017unsupervised} proposed a novel augmentation-based method to generate labelled data with a similar distribution to the target domain for robust speech recognition. A vanilla VAE was trained in an unsupervised way to learn a disentangled latent representation of speech which can be modified for generating expected target domain data. However, the disentangled image attributes in the latent space are a challenging goal to achieve. Instead, we employ a conditional VAE and the domain information can be incorporated and fed into the decoder for target domain sample generation. Volpi et al. \cite{volpi2018adversarial} performed data augmentation in the feature space by devising a feature generator trained with a Conditional Generative Adversarial Network (CGAN). Our approach is similar to this in the sense of feature augmentation whilst we aim to augment data by feature transformation across domains rather than from random noises. Huang et al. \cite{huang2018auggan} proposed GAN based models for image-to-image translation and evaluated the performance in object detection rather than image classification which is our focus in this work. Lv et al. \cite{lv2019targan} also utilised GAN to generate target domain data given class labels to improve the classifier training. Following these studies, in our work, a novel {\it norm-VAE} is proposed to generate target domain samples by feature transformation across domains and its effectiveness is demonstrated through comparative experiments.

\subsection{Variational Autoencoder for UDA}\label{sec:related_vae}

Variational Autoencoder (VAE) has been a prevalent generative model for data generation and it has been used for UDA in literature \cite{hsu2017unsupervised,hou2019unsupervised,wang2019vae,ilse2020diva,xu2020adversarial,chen2019deep}.

Hou et al. \cite{hou2019unsupervised} aim to generate synthetic target-domain data with VAEs trained domain-wisely. Subsequently, the higher-level and lower-level layers of the decoders for source and target domains are cross-stacked to form new VAEs which can be used to transform images from one domain to the other. However, the effectiveness of the idea was only validated on digits data in \cite{hou2019unsupervised} and is questionable for more complicated image classification tasks. In contrast to pixel-level image generation, a more reliable alternative is employed in our work which aims to generate image features with a simplified VAE model. Wang et al.\cite{wang2019vae} also used VAE in the feature space for speech signal representation learning. However, their work focused on the latent code vectors $z$ generated by the encoder of VAE whilst our goal is to generate synthetic features in the original feature space. We also investigated the effect of latent code vectors in our preliminary experiments but did not observe favourable performance enhancement in the image classification tasks.
Chen et al. \cite{chen2019deep} utilized two-stream Wasserstein Autoencoders to map the data from four domains (i.e. real source, real target, synthetic source and synthetic target) into a common subspace towards better classification performance. By contrast, our work also concern data from these four domains whilst the classification is carried out in the original feature space without the need of learning a latent space.

\section{Problem Formulation}\label{sec:problem}
Before presenting our method, we describe the standard problem formulation of UDA for image classification.
Given a labelled dataset $\mathcal{D}^s = \{(\bm{x}^s_i,y^s_i)\}, i = 1,2,...,n_s$ from the source domain $\mathcal{S}$, $\bm{x}^s_i \in \mathbb{R}^{d}$ represents the feature vector of $i$-th labelled sample in the source domain, $d$ is the feature dimension and $y^s_i \in \mathcal{Y}^s$ denotes the corresponding label. UDA aims to classify an unlabelled data set $\mathcal{D}^t = \{\bm{x}^t_i\}, i=1,2,...,n_t$ from the target domain $\mathcal{T}$, where $\bm{x}^t_i \in \mathbb{R}^{d}$ represents the feature vector in the target domain. The target label space $\mathcal{Y}^t$ is equal to the source label space $\mathcal{Y}^s$. It is assumed that both the labelled source domain data $\mathcal{D}^s$ and the unlabelled target domain data $\mathcal{D}^t$ are available for model learning. As a result, most existing UDA approaches are are evaluated in the transductive learning setting. Cases of inductive learning settings where evaluation on new target data that are not accessed during training are also considered in the literature \cite{tzeng2017adversarial,long2018conditional,chen2019progressive,saito2018maximum}. Our proposed methods apply to both settings.

\section{Proposed Method}\label{sec:method}
In this section, we first present a computationally simple approach to UDA for classification problems. The approach is based on the hypothesis {\it a unified classifier for both source and target domains can be trained in the original homogeneous feature space in spite of the domain shift across domains by supervised learning}. Pseudo-labelled target domain data are combined with labelled source domain data to train the unified classifier for both source and target domains.
Subsequently, we describe our proposed generative model \textit{norm-VAE} which is used to generate synthetic features to augment the training data for classifier training.

\subsection{Revisiting Selective Pseudo-Labelling} \label{sec:spl}
We aim to learn a unified classifier $y=f(x;\bm{\theta})$ for both source and target domains. The classifier $f(x;\bm{\theta})$ can be implemented as a shallow CNN model for image classification when the input $x$ are raw images or a linear two-layer Multi-Layer Perceptron (MLP, containing an input layer and an output layer) when the input $x$ are image features.
As the first step, we train the classifier with labelled source domain data. The trained classifier is subsequently used to classify unlabelled target domain samples and get their pseudo labels $y_i^t, i=1,2,...,n_t$. The confidence score $s(y_i^t)$ of the pseudo label $y_i^t$ can also be obtained from the softmax layer of the classifier.
The pseudo-labelled target domain samples are combined with the labelled source domain samples to re-train the classifier so that the classifier can gain the capability of separating target domain samples. The updated classifier is again used to update the pseudo-labels of target domain samples. This process can be repeated for multiple iterations towards an optimal classifier and better classification performance.

One key to the above pseudo-labelling strategy is the selection of pseudo-labelled target domain samples for training in each iteration. Instead of using all the pseudo-labelled target samples for classifier training, it has been proved progressively selecting a fraction of the target domain samples for training is beneficial \cite{chen2019progressive,wang2020unsupervised}. Following the previous works in \cite{wang2019unifying} and \cite{wang2020unsupervised}, we select pseudo-labelled target samples with top confidence scores class-wisely and add them to the training data set in each iteration. Specifically, we consider the pseudo-labels class-wisely and select top-$K$ confident pseudo-labelled target domain samples for each class. 

Distinct from existing selective pseudo-labelling in \cite{wang2019unifying} and \cite{wang2020unsupervised}, the number of selected pseudo-labelled target domain samples $N(c,k)$ for $c$-th class in $k$-th iteration is determined as follows:
\begin{equation}
    \label{eq:topk}
    N(c,k) = \min\{\frac{k}{T} \frac{n_t}{C}, \hat{n}_t(c,k)\}
\end{equation}
where $T$ is the number of iterations empirically set as 10 in our experiments; $n_t$ is the number of target domain samples; $C$ is the number of classes and $\hat{n}_t(c,k)$ denotes the number of target domain samples predicted to be from $c$-th classes in $k$-th iteration.
In contrast, $N(c,k)$ is set as $(k\hat{n}_t(c,k))/T$ in previous work \cite{wang2020unsupervised}. 
That is, the number of selected pseudo-labelled samples $N(c,k)$ is proportional to the number of predicted pseudo-labels $\hat{n}_t(c,k)$ for a specific class. As a result, there can be a large number of selected pseudo-labelled samples for some classes whilst very limited pseudo-labelled samples for other classes.
Our pseudo-label selection strategy indicated in Eq.(\ref{eq:topk}) allows balanced pseudo-labelled target samples across different classes. This naive Selective Pseudo-Labelling (naive-SPL) approach is summarized in Algorithm \ref{alg:naive_SPL}. 

\begin{algorithm}[tb]
	\caption{The method of naive Selective Pseudo-Labelling (naive-SPL)}
	\label{alg:naive_SPL}
	\renewcommand{\algorithmicrequire}{\textbf{Input:}}
	\renewcommand{\algorithmicensure}{\textbf{Output:}}
	\begin{algorithmic}[1]
		\REQUIRE Labelled source data set $\mathcal{D}^s = \{(\bm{x}^s_i,y^s_i)\}, i = 1,2,...,n_s$ and unlabelled target data set $\mathcal{D}^{t}=\{\bm{x}_i^t\},i=1,2,...,n_{t}$, number of iteration $T$.
		\ENSURE A unified classifier $f(x;\bm{\theta})$ and predicted labels $\{\hat{y}^t\}$ for target domain samples.
		\STATE initialise $k=0$;
		\STATE Training the classifier $f(x;\bm{\theta_0})$ using only source data $\mathcal{D}^s$;
		\STATE Assign pseudo labels for all target data;
		\WHILE {$k < T$}
		\STATE $k \leftarrow k+1$;
		\STATE Select a subset of pseudo-labelled target data $\mathcal{S}_k \in \mathcal{\hat{D}}^t $ using Eq. (\ref{eq:topk});
		\STATE Re-training the classifier using $\mathcal{D}^s$ and $\mathcal{S}_k$;
		\STATE Update pseudo labels for all target data.
		\ENDWHILE
	\end{algorithmic}
\end{algorithm}

\begin{figure*}
    \centering
    {\includegraphics[width=\textwidth]{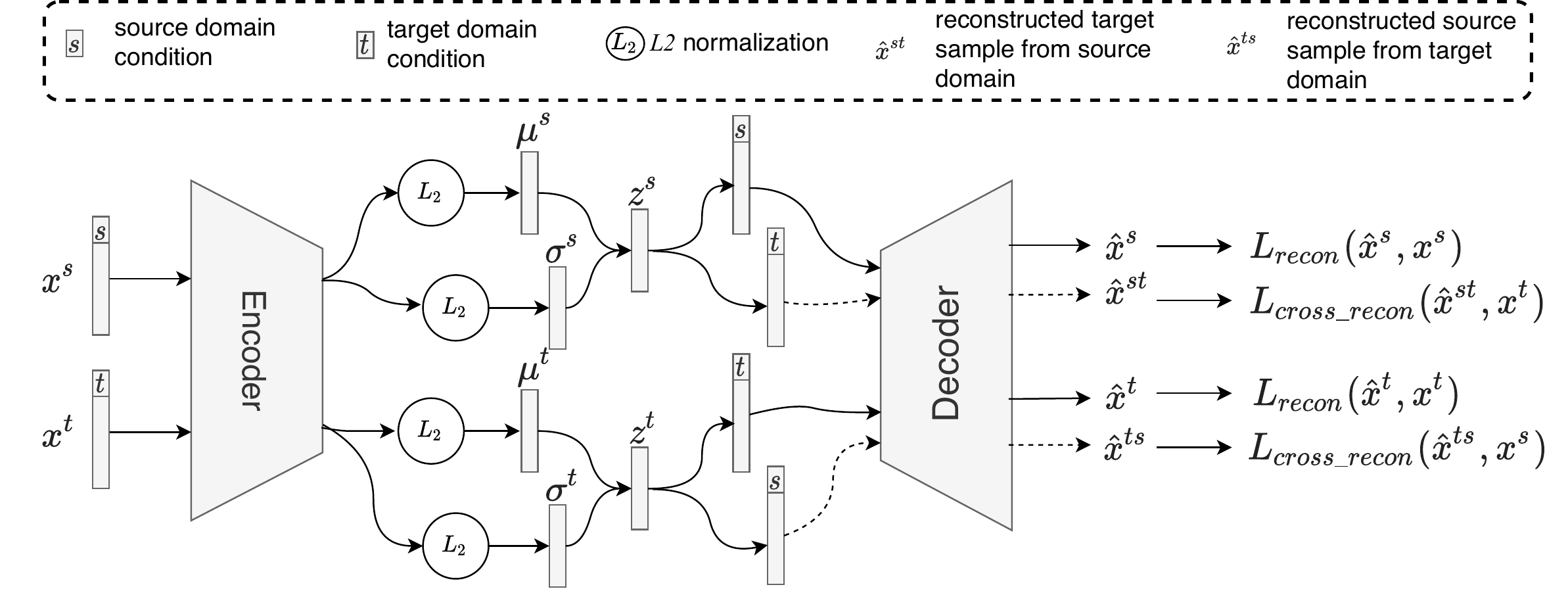}}
    {\caption{The diagram of norm-VAE used for data augmentation. The encoder and decoder are conditioned on the domain label $s$ or $t$. Given a source domain sample $x^s$ as the input, the model generates reconstructed samples $\hat{x}^s$ and $\hat{x}^{st}$ in the source and target domains respectively. Similarly, the model can take a target domain sample $x^t$ as the input and generates $\hat{x}^t$ and $\hat{x}^{ts}$.}
        \label{fig:framework}}
\end{figure*}

\subsection{Data Augmentation Using norm-VAE} \label{sec:normvae}
As opposed to the existing methods of UDA, our proposed {\it naive-SPL} does not aim to explicitly address the distribution discrepancy. Instead, it focuses on the issue of training data scarcity. Following this direction, we propose a novel norm-VAE model to further address the training data scarcity issue in the target domain by generating synthetic target domain features from labelled source domain ones. 

Our proposed generative model is based on conditional VAE (CVAE) \cite{sohn2015learning} and is conditioned on domain labels rather than class labels. As illustrated in Figure \ref{fig:framework}, given an input sample $x$ from the source or target domain, the encoder aims to learn a distribution $p_{\theta}(\bm{z})$ (approximated by $q_{\Phi}(\bm{z}|\bm{x},d)$) from which the latent encoding vector $\bm{z}$ can be sampled and subsequently fed into the decoder to reconstruct the input feature $\hat{\bm{x}}$, where $d$ denotes the domain label condition (i.e. $d\in \{s,t\}$).  The decoder can be parameterized by $p_{\theta}(\bm{x}|\bm{z},d)$. As a result, the model is expected to generate synthetic target domain samples from those in the source domain and vice versa. To this end, we make some essential modifications to the traditional CVAE in two aspects: replacing the Kullback-Leibler divergence regularization by $L_2$ normalization and training the model using paired source and target domain samples.

In traditional CVAE, the loss function is composed of two components as follows:
\begin{equation}
    \label{eq:loss_cvae}
    \begin{aligned}
    \mathcal{L}_{CVAE}(\Phi,\theta;\bm{x}) = & \mathcal{L}_{recon} \big(\bm{x},\bm{\hat{x}}) \\ 
& +  D_{KL}\big(\mathcal{N}(\mu_{\bm{x}},\sigma_{\bm{x}}) || \mathcal{N}(\bm{0},\bm{I})\big) 
    \end{aligned}
\end{equation}
where the first terms represents the reconstruction error and the second term is the KL-divergence between the learned distribution and the standard Normal distribution. The KL-divergence is a regularization term forcing the learned latent codes $\bm{z}$ to follow the standard Normal distribution. This regularization enables the learned model to gain the capability of generating meaningful data from a random latent code $\bm{z}$ sampled from the standard Normal distribution. However, our aim is to generate data from latent codes which are encoded from real input data rather than randomly sampled latent codes, To this end, the latent codes need to be class discriminative and hence we relax the constraint of $D_{KL}$ in Eq.(\ref{eq:loss_cvae}) by replacing it with $L_2$ normalization applied to the outputs of the encoder $\mu$ and $\sigma$. $L_2$ normalization forces the mean and variance vectors located on the hyper-sphere rather than the neighbourhood of the origin to promote the class discriminative property of latent codes $\bm{z}$.

To enable the capability of generating synthetic data across domains, we train the norm-VAE in a novel way. Specifically, we use paired data $\{\bm{x}^s,\bm{x}^t\}$ from source and target domains which belong to the same class. The class information for unlabelled target domain data can be obtained by pseudo-labelling as described in the previous section. The paired data are fed into the norm-VAE and a set of reconstructions are generated as $\{\hat{\bm{x}^s}, \hat{\bm{x}^{st}}, \hat{\bm{x}^t}, \hat{\bm{x}^{ts}}\}$ (c.f. Figure \ref{fig:framework}). The loss function is formulated as:
\begin{equation}
    \label{eq:loss_normvae}
    \begin{aligned}
    & \mathcal{L}_{norm-VAE}(\Phi,\theta;\bm{x}) =  (\mathcal{L}_{recon} \big(\bm{x^s},\bm{\hat{x}^s}) + \mathcal{L}_{recon} (\bm{x^t},\bm{\hat{x}^t})\big) \\ 
&+  \big(\mathcal{L}_{cross\_recon} (\bm{x^s},\bm{\hat{x}^{ts}}) + \mathcal{L}_{cross\_recon} (\bm{x^t},\bm{\hat{x}^{st}})\big)
    \end{aligned}
\end{equation}

The first two terms measure the reconstruction errors for source and target domain samples respectively. The last two terms are cross-domain reconstruction errors. Although the samples in the pair of $\{\bm{x^s},\bm{\hat{x}^{ts}}\}$ or $\{\bm{x^t},\bm{\hat{x}^{st}}\}$ are from the same class, they are not necessarily two views of the same image. To reduce the cross-domain reconstruction errors, the encoder has to preserve class information in the latent code space. As a result, the use of cross-domain reconstruction loss $\mathcal{L}_{cross\_recon}$ facilitates the model to generate class discriminative synthetic data across domains. 

The norm-VAE model is incorporated into the selective pseudo-labelling framework described in the previous section so that the classifier training can be enhanced by combining the real training data and synthetic data generated by norm-VAE. The norm-VAE is trained with labelled source domain data and pseudo-labelled target domain data in each iteration. The method of our proposed norm-VAE-SPL is summarized in Algorithm \ref{alg:norm_VAE_SPL} where the differences from Algorithm \ref{alg:naive_SPL} are highlighted in {\bf bold}.

\begin{algorithm}[tb]
	\caption{The method of SPL with data augmentation by norm-VAE (norm-VAE-SPL)}
	\label{alg:norm_VAE_SPL}
	\renewcommand{\algorithmicrequire}{\textbf{Input:}}
	\renewcommand{\algorithmicensure}{\textbf{Output:}}
	\begin{algorithmic}[1]
		\REQUIRE Labelled source data set $\mathcal{D}^s = \{(\bm{x}^s_i,y^s_i)\}, i = 1,2,...,n_s$ and unlabelled target data set $\mathcal{D}^{t}=\{\bm{x}_i^t\},i=1,2,...,n_{t}$, number of iteration $T$.
		\ENSURE A unified classifier $f(x;\bm{\theta})$ and predicted labels $\{\hat{y}^t\}$ for target domain samples.
		\STATE initialise $k=0$;
		\STATE Training the classifier $f(x;\bm{\theta_0})$ using only source data $\mathcal{D}^s$;
		\STATE Assign pseudo labels for all target data;
		\WHILE {$k < T$}
		\STATE $k \leftarrow k+1$;
		\STATE Select a subset of pseudo-labelled target data $\mathcal{S}_k \in \mathcal{\hat{D}}^t $ using Eq. (\ref{eq:topk});
		\STATE {\bf Training the norm-VAE model using  $\mathcal{D}^s$ and $\mathcal{S}_k$ by minimizing the loss in Eq.(\ref{eq:loss_normvae});}
		\STATE Re-training the classifier using real data from $\mathcal{D}^s$ and $\mathcal{S}_k$, {\bf and their corresponding synthetic data generated by norm-VAE};
		\STATE Update pseudo labels for all target data.
		\ENDWHILE
	\end{algorithmic}
\end{algorithm}

\subsection{Model Architectures and Computational Complexity} \label{sec:method_complexity}
The computational cost of Algorithm \ref{alg:naive_SPL} depends on the classifier training itself and the number of iterations $T$. In our experiments, we use a CNN architecture from \cite{saito2018maximum} as the classifier which consists of two convolutional layers and three fully-connected layers for digit classification. For image classification datasets, we use deep features (i.e. ResNet50) and a linear MLP consisting of only the input and output layers which is computationally efficient.

The method {\it norm-VAE-SPL} in Algorithm \ref{alg:norm_VAE_SPL} involves one additional step of training the {\it norm-VAE} model. The encoder and decoder are implemented as 3-layer MLP (i.e. $d^{x} \to 512 \to d^{z}$ and $d^{z} \to 512 \to d^{x}$) with ReLU layers and a dropout rate of 0.5 applied on the intermediate activations. $d^{x}$ and $d^{z}$ are the dimensionality of input $\bm{x}$ and latent code $\bm{z}$ respectively. The value of $d^z$ is set as 64 in our experiments.

\begin{figure*}
	\centering
	{\includegraphics[width=1\textwidth]{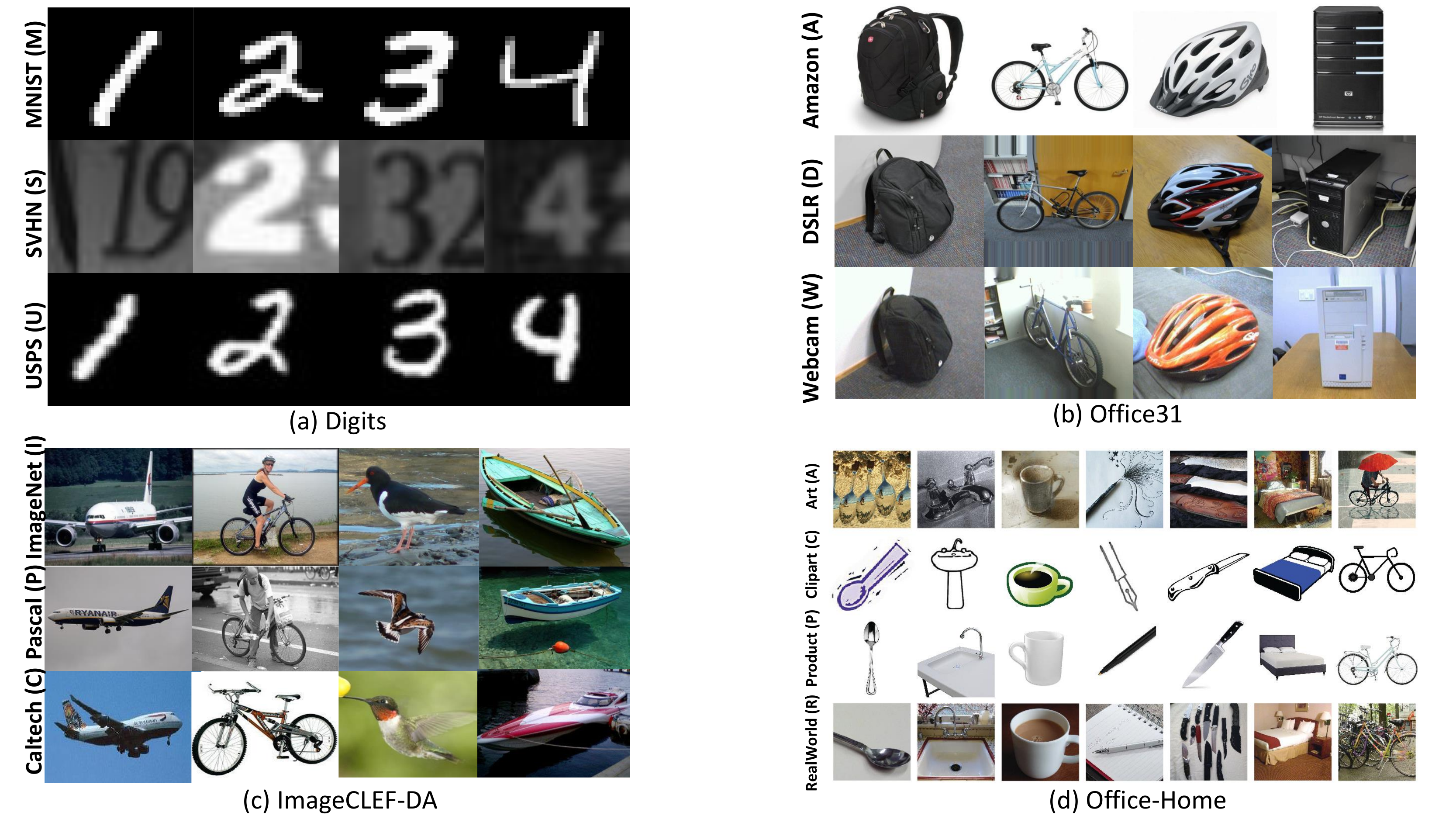}}
	{\caption{Exemplar images from different domains of four datasets used in our experiments (The Office-Caltech dataset consists of the same domains as Office31 and one additional Caltech domain; exemplar images for the Office-Home dataset (d) originate from \cite{venkateswara2017deep}; best viewed in color).}
		\label{fig:datasets}}
\end{figure*}

\section{Experiments and Results}\label{sec:experiments}

\begin{table}[!t]
	\centering
	{
		\centering
		\caption[]{Classification Accuracy (\%) of UDA on Digits dataset (M: MNIST, U: USPS, S: SVHN).
		}
		\label{table:uda_digit}
		\resizebox{1\columnwidth}{!}{%
			\begin{tabular}{lcccc}
				\hline
				Method & M$\to$ U & U $\to$ M & M $\to$ S & Average\\ \hline
				ADDA \cite{tzeng2017adversarial} & 89.4 & 90.1 & 76.0 & 85.2\\
				GTA \cite{sankaranarayanan2018generate} & 95.3 & 90.8 & 92.4 & 92.8\\
				MCD \cite{saito2018maximum} & \underline{96.5} & 94.1 & 96.2 & 95.6\\
				MCD+CAT \cite{deng2019cluster} & 96.3 & 95.2 & 97.1 & 96.3\\
				\footnotesize{rRevGrad+CAT \cite{deng2019cluster}} & 94.0 & 96.0 & \underline{98.8} & 96.3\\
				SHOT \cite{liang2020we} & \bf 98.0 & \bf 98.4 & \bf 98.9 & \bf 98.4\\
				\hline
				Baseline & 30.1 & 51.3 & 83.1 & 54.8\\
				naive-SPL* & 88.2 & 91.7 & 90.7 & 90.2 \\
				naive-SPL (Ours) & 95.8 & \underline{97.7} & 98.0 & \underline{97.2}\\
				norm-VAE-SPL (Ours) & 95.8 & \underline{97.7} & 98.0 & \underline{97.2}\\
				\hline
			\end{tabular}%
		}
	}
\end{table}


\begin{table*}[!t]
	\centering
	{
		\centering
		\caption[]{Classification Accuracy (\%) on Office-Caltech dataset using Decaf6 features. Each column displays the results of a pair of source $\to$ target setting.}
		\label{table:uda_o10}
		\resizebox{\columnwidth}{!}{%
			\begin{tabular}{lccccccccccccc}
				\hline
				Method & C$\to$A & C$\to$W & C$\to$D & A$\to$C&A$\to$W & A$\to$D&W$\to$C & W$\to$A & W$\to$D & D$\to$C & D$\to$A & D$\to$W & Average \\ \hline
				DDC\cite{tzeng2014deep}  & 91.9 & 85.4 & 88.8 & 85.0 & 86.1 & 89.0 & 78.0 & 84.9 & \textbf{100.0} & 81.1 & 89.5 & 98.2 & 88.2\\
				DAN\cite{long2015learning}  & 92.0 & 90.6 & 89.3 & 84.1 & 91.8 & 91.7 & 81.2 & 92.1 & \textbf{100.0} & 80.3 & 90.0 & 98.5 & 90.1\\
				DCORAL\cite{sun2016deep} & 92.4 & 91.1 & 91.4 & 84.7& - & - & 79.3 & - & - & 82.8 & - & - & -\\
				CORAL\cite{sun2017correlation}& 92.0 & 80.0 & 84.7 & 83.2 & 74.6 & 84.1 & 75.5 & 81.2 & \textbf{100.0} & 76.8 & 85.5 & 99.3 & 84.7\\
				SCA\cite{ghifary2017scatter}  & 89.5 & 85.4 & 87.9 & 78.8 & 75.9 & 85.4 & 74.8 & 86.1 & \textbf{100.0} & 78.1 & 90.0 & 98.6 & 85.9 \\
				JGSA\cite{zhang2017joint} & 91.4 & 86.8 & 93.6 & 84.9 & 81.0 & 88.5 & 85.0 & 90.7 & \textbf{100.0} & 86.2 & 92.0 & \underline{99.7} & 90.0\\
				MEDA\cite{wang2018visual} & 93.4 & \underline{95.6} & 91.1 & 87.4 & 88.1 & 88.1 & \textbf{93.2} & \textbf{99.4} & \underline{99.4} & 87.5 & \underline{93.2} & 97.6 & 92.8\\
				CAPLS \cite{wang2019unifying} & 90.8 & 85.4 & \underline{95.5} & 86.1 & 87.1 & \textbf{94.9} & 88.2 & 92.3 & \textbf{100.0} & \underline{88.8} & 93.0 & \textbf{100.0}& 91.8\\
				SPL \cite{wang2020unsupervised} &92.7 & 93.2 & \textbf{98.7} &87.4 & 95.3 & 89.2 & 87.0 & 92.0 & \textbf{100.0} & 88.6 & 92.9 & 98.6 & \underline{93.0}\\
				\hline
				Baseline & 91.8 & 80.5 & 87.5 & 84.7 & 76.1 & 84.3 & 74.1 & 78.3 & \bf 100.0 & 77.3 & 85.3 & 98.3 & 84.9 \\
				naive-SPL* & 92.6 & 89.2 & 95.4 & \underline{87.5} & 89.5 & \underline{93.0} & 87.9 & 91.9 & \bf 100.0 & \bf 89.1 & 92.9 & 99.3 & 92.4 \\
				naive-SPL (Ours) & \bf 94.1 & 92.9 & 88.5 & 86.9 & \underline{95.6} & 91.3 & 87.5 & \underline{94.1} & \bf 100.0 & 88.7 & \bf{94.1} & 99.3 & 92.8 \\
				norm-VAE-SPL (Ours) & \underline{94.0} & \bf 97.6 & 90.8 & \bf 88.1 & \bf 97.3 & 92.0 & \underline{88.4} & 93.0 & \underline{99.4} & 87.9 & 93.0 & 99.3 & \bf 93.4 \\
				\hline
			\end{tabular}%
		}
	}
\end{table*}

\begin{table}[!t]
	\centering
	{
		\centering
		\caption[]{Classification Accuracy (\%) on Office31 dataset using either ResNet50 features or ResNet50 based deep models.
		}
		\label{table:uda_o31}
		\resizebox{1\columnwidth}{!}{%
			\begin{tabular}{lccccccc}
				\hline
				Method & \scriptsize{A$\to$W} & \scriptsize{D$\to$W} & \scriptsize{W$\to$D} & \scriptsize{A$\to$D} & \scriptsize{D$\to$A} & \scriptsize{W$\to$A} & Avg \\ \hline
				RTN\cite{long2016unsupervised} & 84.5 & 96.8 & 99.4 & 77.5 & 66.2 & 64.8 & 81.6\\
				MADA\cite{pei2018multi} & 90.0 & 97.4 & 99.6 & 87.8 & 70.3 & 66.4 & 85.2 \\
				MEDA\cite{wang2018visual} & 86.2 & 97.2 & 99.4 & 85.3 & 72.4 & 74.0 & 85.7\\
				GTA \scriptsize{\cite{sankaranarayanan2017generate}} & 89.5& 97.9 & 99.8& 87.7 & 72.8 & 71.4& 86.5\\
				iCAN\cite{zhang2018collaborative} & 92.5 & \underline{98.8} & \textbf{100.0} & 90.1 & 72.1 & 69.9 & 87.2 \\
				CDAN-E\cite{long2018conditional} & \underline{94.1} & 98.6 & \textbf{100.0} & 92.9 & 71.0 & 69.3 & 87.7\\
				JDDA\cite{chen2018joint} & 82.6 & 95.2 & 99.7 & 79.8 & 57.4 & 66.7 & 80.2\\
				SymNets\cite{zhang2019domain} & 90.8 & \underline{98.8} & \textbf{100.0} & \textbf{93.9} & 74.6 & 72.5 & \underline{88.4}\\
				TADA \cite{wang2019transferable} &  \textbf{94.3} & 98.7 & 99.8 & 91.6 & 72.9 & 73.0 & \underline{88.4}\\

				CAPLS \scriptsize{\cite{wang2019unifying}} & 90.6 & 98.6 & 99.6 & 88.6 & \underline{75.4} & \underline{76.3} & {88.2}\\
				SPL \cite{wang2020unsupervised} & 92.7 & 98.7 & 99.8 & \underline{93.0} & \textbf{76.4} & \textbf{76.8} & \textbf{89.6} \\
				\hline
				Baseline & 73.3 & 97.5 & 99.6 & 75.8 & 68.0 & 67.6 & 80.3 \\
				naive-SPL* & 84.6 & \bf 98.9 & 99.8 & 81.4 & 70.4 & 70.6 & 84.3 \\
				naive-SPL (Ours) & 88.6 & 98.1 & \underline{99.9} & 82.0 & 73.6 & 73.4 & 85.9 \\
				norm-VAE-SPL (Ours) & 92.9 & 96.9 & 98.7 & 88.8 & 74.5 & 74.1 & 87.6 \\
				\hline
			\end{tabular}%
		}
	}
\end{table}

\begin{table}[!t]
	\centering
	{
		\centering
		\caption[]{Classification Accuracy (\%) on ImageCLEF-DA dataset using either ResNet50 features or ResNet50 based deep models.
		}
		\label{table:uda_clef}
		\resizebox{1\columnwidth}{!}{%
			\begin{tabular}{lccccccc}
				\hline
				Method &I$\to$P & P$\to$I & I$\to$C & C$\to$I & C$\to$P & P$\to$C & Avg \\ \hline
				RTN\cite{long2016unsupervised} & 75.6 & 86.8 & 95.3 & 86.9 & 72.7 & 92.2 & 84.9\\
				MADA\cite{pei2018multi} & 75.0 & 87.9 & 96.0 & 88.8 & 75.2 & 92.2 & 85.8 \\
				iCAN\cite{zhang2018collaborative} & 79.5 & 89.7 & 94.7 & 89.9 & 78.5 & 92.0 & 87.4 \\
				CDAN-E\cite{long2018conditional} & 77.7 & 90.7 & \textbf{97.7} & 91.3 & 74.2 & 94.3 & 87.7\\
				SymNets\cite{zhang2019domain} & \underline{80.2} & 93.6 & \underline{97.0} & 93.4 & 78.7 & \textbf{96.4} & 89.9\\

				MEDA\cite{wang2018visual} & 79.7 & 92.5 & 95.7 & 92.2 & 78.5 & 95.5 & 89.0\\
				SPL \cite{wang2020unsupervised} & 78.3 & \textbf{94.5} & 96.7 & \textbf{95.7} & \textbf{80.5} & \underline{96.3} & \underline{90.3}\\
				\hline
				Baseline& 79.1 & 89.2 & 94.0 & 86.0 & 69.9 & 92.4 & 85.1 \\
				naive-SPL* & 77.0 & 90.7 & 95.9 & 92.1 & 73.4 & 93.7 & 87.1 \\
				naive-SPL (Ours) & 80.0 & 91.5 & 96.2 & 94.3 & 79.1 & 95.6 & 89.4 \\
				norm-VAE-SPL (Ours) & \bf 80.3 & \underline{93.9} & 96.9 & \underline{94.6} & \underline{80.4} & \underline{96.3} & \bf 90.4 \\
				\hline
			\end{tabular}%
		}
	}
\end{table}

In this section, we describe our experiments on commonly used datasets for unsupervised domain adaptation for image classification (i.e. Digits, Office-Caltech \cite{gong2012geodesic}, Office31 \cite{saenko2010adapting}, ImageCLEF-DA \cite{caputo2014imageclef} and Office-Home \cite{venkateswara2017deep}). Our approach is firstly compared with state-of-the-art UDA approaches to evaluate its effectiveness. An ablation study is conducted to demonstrate the effects of different components and hyper-parameters in our approach. Finally, we investigate how different hyper-parameters affect the performance.

\subsection{Datasets}\label{sec:dataset}

To make a thorough evaluation, we conduct experiments on five commonly used datasets including one digit classification dataset and four image classification datasets. Exemplar images from different domains are shown in Figure \ref{fig:datasets} for four datasets. The Office-Caltech dataset is not shown since it consists of the same 3 domains as those in Office31 and the Caltech domain in ImageCLEF-DA. More details of these datasets are described as follows.

\textbf{Digit} classification is a commonly used benchmark for unsupervised domain adaptation. We follow existing works \cite{tzeng2017adversarial,long2018conditional,chen2019progressive,saito2018maximum} and consider three domain adaptation tasks (i.e. MNIST $\to$ USPS, USPS $\to$ MNIST and MNIST $\to$ SVHN) on three digit datasets: MNIST, USPS and SVHN. There are 60,000/10,000 images for training/testing in MNIST, 7,291/2,007 in USPS, and 73,257/26,032 in SVHN. In each dataset, there are 10 classes of digit 0--9. 

\textbf{Office-Caltech} \cite{gong2012geodesic} consists of four domains: Amazon (A, images downloaded from online merchants), Webcam (W, low-resolution images by a web camera), DSLR (D, high-resolution images by a digital SLR camera) and Caltech-256 (C).  Ten common classes from all four domains are used: backpack, bike, calculator, headphone, computer-keyboard, laptop-101, computer-monitor, computer-mouse, coffee-mug, and video-projector. There are 2533 images in total with 8 to 151 images per category per domain.

\textbf{Office31} \cite{saenko2010adapting} consists of three domains: Amazon (A), Webcam (W) and DSLR (D). There are 31 common classes for all three domains containing 4,110 images in total. 

\textbf{ImageCLEF-DA} \cite{caputo2014imageclef} consists of four domains. We follow the existing works \cite{zhang2019domain} using three of them in our experiments: Caltech-256 (C), ImageNet ILSVRC 2012 (I), and Pascal VOC 2012 (P). There are 12 classes and 50 images for each class in each domain. 

\textbf{Office-Home} \cite{venkateswara2017deep} is another dataset recently released for evaluation of domain adaptation algorithms. It consists of four different domains: Artistic images (A), Clipart (C), Product images (P) and Real-World images (R). There are 65 object classes in each domain with a total number of 15,588 images.

\subsection{Experimental Setting}\label{sec:setting}
The algorithm is implemented in PyTorch\footnote{Code is available: https://github.com/hellowangqian/UDA-norm-VAE}. For digit classification, we use the same CNN model designed by \cite{saito2018maximum}. In each domain adaptation task, the labelled training data from the source domain and the unlabelled training data from the target domain are used to train the classifier which is subsequently evaluated on the test data from the target domain. 
For the Office-Caltech dataset, we use deep features Decaf6 \cite{donahue2014decaf}  (activations of the 6\textit{th} fully connected layer of a convolutional neural network trained on ImageNet, $d=4096$) which were commonly used in existing works for a fair comparison with the state of the arts.
For the other three datasets, ResNet50 \cite{he2016deep} features ($d=2048$) are used in our experiments to allow a direct comparison with other methods.

\begin{table*}[!htbp]
	\centering
	{
		\centering
		\caption[]{Classification Accuracy (\%) on Office-Home dataset using either ResNet50 features or ResNet50 based deep models.}
		\label{table:uda_o65}
		\resizebox{\columnwidth}{!}{%
			\begin{tabular}{lccccccccccccc}
				\hline
				Method & A$\to$C & A$\to$P & A$\to$R & C$\to$A&C$\to$P & C$\to$R&P$\to$A & P$\to$C & P$\to$R & R$\to$A & R$\to$C & R$\to$P & Average \\ \hline
				JAN\cite{long2017deep} & 45.9 & 61.2 & 68.9 & 50.4 & 59.7 & 61.0 & 45.8 & 43.4 & 70.3 & 63.9 & 52.4 & 76.8 & 58.3\\
				CDAN-E \cite{long2018conditional} & 50.7 & 70.6 & 76.0 & 57.6 & 70.0 & 70.0 & 57.4 & 50.9 & 77.3 & 70.9 & 56.7 & 81.6 & 65.8\\
				MEDA\cite{wang2018visual} & \underline{54.6} & 75.2 & 77.0 & 56.5 & 72.8 & 72.3 & 59.0 & 51.9 & 78.2 & 67.7 & \underline{57.2} & 81.8 & 67.0\\
				SymNets \cite{zhang2019domain} & 47.7 & 72.9 & 78.5 & 64.2 & 71.3 & 74.2 & 64.2 & 48.8 & 79.5 & \textbf{74.5} & 52.6 & 82.7 & 67.6 \\ 
				TADA \cite{wang2019transferable} & 53.1 & 72.3 & 77.2 & 59.1 & 71.2 & 72.1 & 59.7 & \underline{53.1} & 78.4 & \underline{72.4} & \textbf{60.0} & 82.9 & 67.6 \\

				CAPLS \cite{wang2019unifying} & \textbf{56.2} & \textbf{78.3} & 80.2 & \textbf{66.0} & 75.4 & \underline{78.4} & \textbf{66.4} & \textbf{53.2} & \underline{81.1} & {71.6} & 56.1 & \underline{84.3} & \underline{70.6}\\
				SPL \cite{wang2020unsupervised} & 54.5 & \underline{77.8} & \textbf{81.9} & \underline{65.1} & \textbf{78.0} & \textbf{81.1} & \underline{66.0} & \underline{53.1} & \textbf{82.8} & 69.9 & 55.3 & \textbf{86.0} & \textbf{71.0} \\
				\hline
				Baseline& 43.1 & 65.1 & 73.7 & 50.7 & 64.4 & 64.6 & 53.8 & 41.6 & 73.8 & 62.7 & 44.3 & 77.4 & 59.6 \\
				naive-SPL*& 44.1 & 72.5 & 77.7 & 52.0 & 70.0 & 70.5 & 52.4 & 39.4 & 77.3 & 62.3 & 44.3 & 79.4 & 61.8 \\
				naive-SPL (Ours) & 52.0 & 74.2 & 79.1 & 56.1 & 74.4 & 74.1 & 56.8 & 49.0 & 78.1 & 61.4 & 52.4 & 80.5 & 65.7 \\
				norm-VAE-SPL (Ours) & 52.7 & 75.6 & \underline{80.9} & 60.4 & \underline{76.1} & 77.0 & 61.8 & 49.8 & 80.6 & 65.4 & 52.8 & 81.8 & 67.9 \\
				\hline
			\end{tabular}%
		}
	}
\end{table*}
\subsection{Comparison with State-of-the-Art Approaches}\label{sec:exp_uda}
We compare our approaches with the most competitive methods including those based on deep features (extracted using deep models such as ResNet50 pre-trained on ImageNet) and deep learning models using pre-trained ResNet50 as the backbones. The classification accuracy of our approaches and the comparative ones are shown in Tables \ref{table:uda_digit}-\ref{table:uda_o65} in terms of each combination of ``source" $\rightarrow$ ``target" domains and the average accuracy over all different combinations. The classification accuracy is calculated as the number of correctly predicted samples over the total number of test samples. For all experiments in this section, each task is repeated for five times with random seeds set as 0-4 to calculate the mean accuracy for this task. We use bold and underlined fonts to indicate the best and the second-best results respectively in each column of the tables.

Our approaches without and with data augmentation are denoted as \textit{\textbf{naive-SPL}} and \textit{\textbf{norm-VAE-SPL}}, respectively. Besides, we conduct an ablation study to investigate the effect of different pseudo-label selection strategies. For this purpose, we consider two more related methods in our experiments. One is denoted as \textit{\textbf{Baseline}} which uses all pseudo-labelled target domain samples without selection for classifier training. The other dubbed as \textit{\textbf{naive-SPL*}} is adapted from our proposed \textit{\textbf{naive-SPL}} by replacing the pseudo-label selection strategy in Eq.(\ref{eq:topk}) with that used in \cite{wang2020unsupervised}.

Table \ref{table:uda_digit} shows the performance of different UDA approaches on the Digits dataset. The classifier is implemented by a shallow CNN model (c.f. Section \ref{sec:method_complexity}) and is trained on raw images. Our proposed {\it naive-SPL} achieves an average accuracy of 97.2\% over three commonly used domain adaptation tasks which is better than all the comparative methods except SHOT \cite{liang2020we} with an average accuracy of 98.4\%. The method {\it norm-VAE-SPL} in this experiment is based on the features extracted by the classifier trained for {\it naive-SPL}. As we can see, there is no performance improvement when data augmentation by norm-VAE is employed in this case. This is because our selective pseudo-labelling strategy described in Section \ref{sec:spl} enables the CNN model to learn domain-invariant features from source and target domains. As a result, the domain shift between the source- and target- domain features used to train the {\it norm-VAE} model is negligible hence the synthetic features generated by {\it norm-VAE} can not provide additional information for classifier training. 

In many real-world applications, however, much deeper and more complicated CNN models than the one used for digit classification are required to extract image features. Training large CNN models on both source and target domain data can be computational expensive and unnecessary \cite{wang2020unsupervised}. In the following experiments on real-world image datasets, we use pre-trained (on ImageNet only) ResNet50 as the feature extractor to extract features of source and target domain images. As shown in Figure \ref{fig:visualization}, a data distribution shift can be clearly observed between source and target domain data. In these cases, our proposed approach including the data augmentation model {\it norm-VAE} demonstrates its effectiveness in improving the classification accuracy as described in the following paragraphs.

Tables \ref{table:uda_o10}-\ref{table:uda_o65} demonstrate the results on image classification datasets. Our proposed {\it naive-SPL} can already achieve quite good performance with the average accuracy of 92.8\% on Office-Caltech, 85.9\% on Office31, 89.4\% on ImageCLEF-DA and 65.7\% on Office-Home. These results are comparable with many complex UDA approaches, especially on Office-Caltech and ImageCLEF-DA datasets. These validates the effectiveness of our proposed selective pseudo-labelling strategy since {\it naive-SPL} is no more than a simple classifier trained with labelled source and pseudo-labelled target domain samples in an iterative manner. With the use of our proposed data augmentation method, {\it norm-VAE-SPL} improves the performance consistently on all of four image classification datasets. As a result, our proposed {\it norm-VAE-SPL} achieves the best average accuracy of 93.4\% and 90.4\% on Office-Caltech and ImageCLEF-DA datasets, respectively. On the other two datasets, {\it norm-VAE-SPL} also performs comparably well with most approaches except CAPLS \cite{wang2019unifying} and SPL \cite{wang2020unsupervised}. It is noteworthy that both of them employ the dimensionality reduction algorithm Locality Preserving Projection (LPP) to learn a latent subspace where source- and target-domain data can be well aligned. These methods require to solve the eignvalue problems and the computational cost is subject to the number of samples in both domains. As a result, they are not suitable for large-scale applications whilst our proposed method does not have such constraints. Besides, our method is intrinsically different from those based on the domain alignment in that we assume a unified classifier can be learned in the original homogeneous feature space in spite of the existence of the domain shift across source and target domains.

We conduct an ablation study by comparing the performance of {\it Baseline}, {\it naive-SPL*} and {\it naive-SPL} and some consistent conclusions can be drawn from this ablation study. Firstly, the {\it Baseline} method using all pseudo-labelled target-domain data without selection is always inferior to the other two methods with selective pseudo-labelling. Specifically, the performance gaps between the {\it Baseline} and {\it naive-SPL*} methods in terms of the average classification accuracy are 7.5\%, 4.0\%, 2.0\% and 2.2\% on Office-Caltech, Office31, ImageCLEF-DA and Office-Home datasets respectively. These results demonstrate that selecting the most confident pseudo-labels progressively is of vital importance to the classifier training. On the other hand, the pseudo-label selection strategy used in \cite{wang2020unsupervised} is inferior to the proposed alternative described in Section \ref{sec:spl} as {\it naive-SPL} outperforms {\it naive-SPL*} by margins of 0.4\%, 1.6\%, 2.3\% and 3.9\% on four image classification datasets respectively. 
\begin{figure*}
	\centering
	{\includegraphics[width=\textwidth]{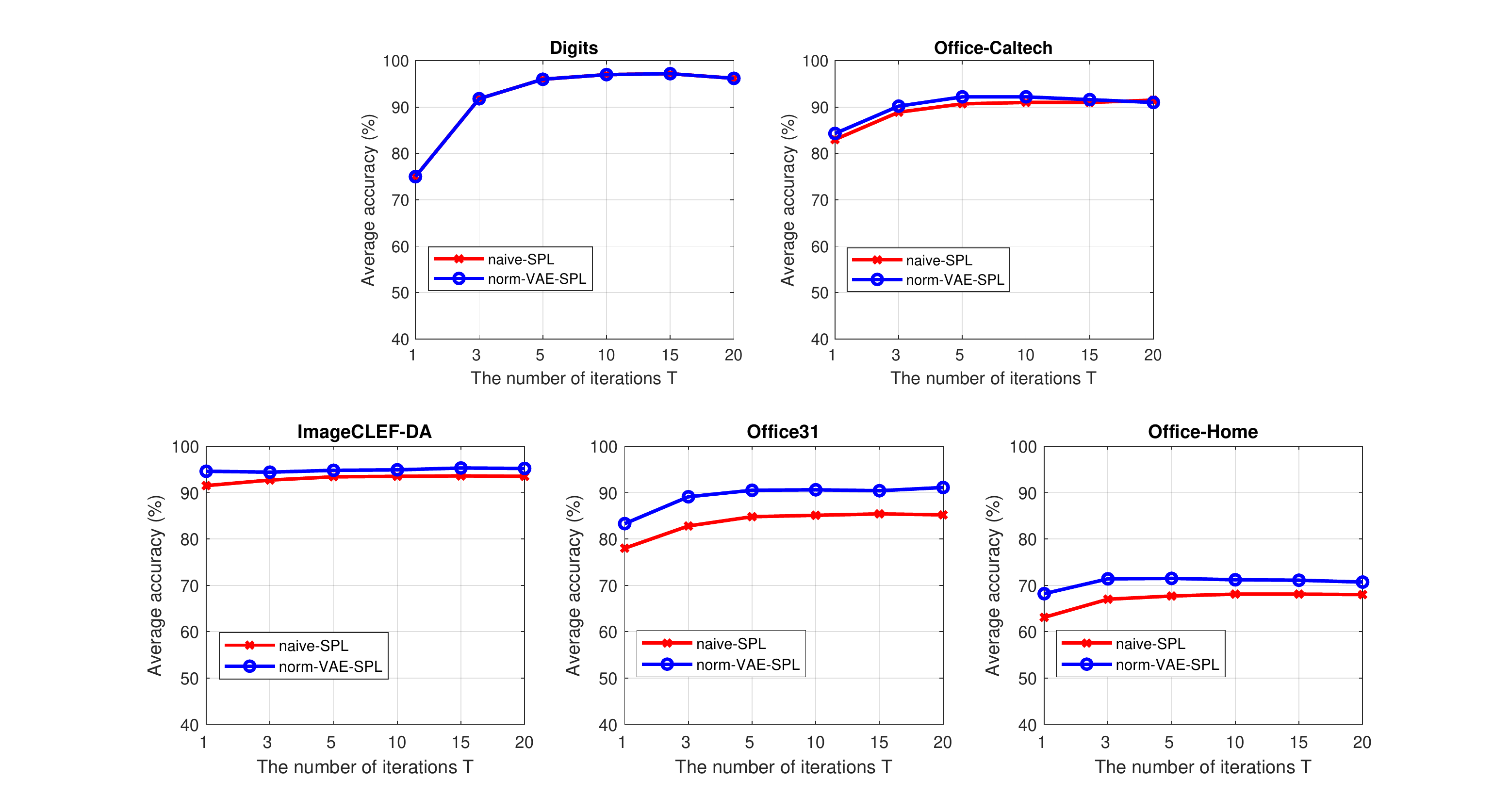}}
	{\caption{Effect of the number of iterations $T$.}
		\label{fig:hyper-parameters}}
\end{figure*}

\subsection{Effects of the Hyper-parameter $T$}
In Algorithms \ref{alg:naive_SPL} and \ref{alg:norm_VAE_SPL}, the number of iterations $T$ is a hyper-parameter which was set as 10 throughout our main experiments. In this experiment, we investigate how the value of $T$ affects the performance of {\it naive-SPL} and {\it norm-VAE-SPL}. To this end, we set the value of $T$ to be 1, 3, 5, 10, 15, 20 respectively and calculate the average accuracy over some representative domain adaptation tasks. Specifically, we consider all three tasks for Digits, three tasks $C\to A/D/W$ for Office-Caltech, two tasks $A\to W/D$ for Office31, two tasks $P \to I/C$ for ImageCLEF-DA and three tasks $A \to C/P/R$ for Office-Home. Each task is repeated for three times with random seeds set as 0, 1 and 2. 

The results are shown in Figure \ref{fig:hyper-parameters} in which the average accuracy over considered tasks are reported for five datasets. As we can see, the number of iterations $T$ has a negligible effect on the performance when it is greater than 5 for both {\it naive-SPL} and {\it norm-VAE-SPL}. For the Digits dataset, significant performance improvement can be observed when $T$ increases from 1 to 5 whilst for other image classification dataset, the optimal value of $T$ varies from 1, 3 to 5 with subtle differences. To summarize, our approaches are not sensitive to the hyper-parameters and perform well enough with a relatively small number of iterations.

\subsection{Data Visualization}\label{sec:visual}
\begin{figure*}
	\centering
	{\includegraphics[width=1\textwidth]{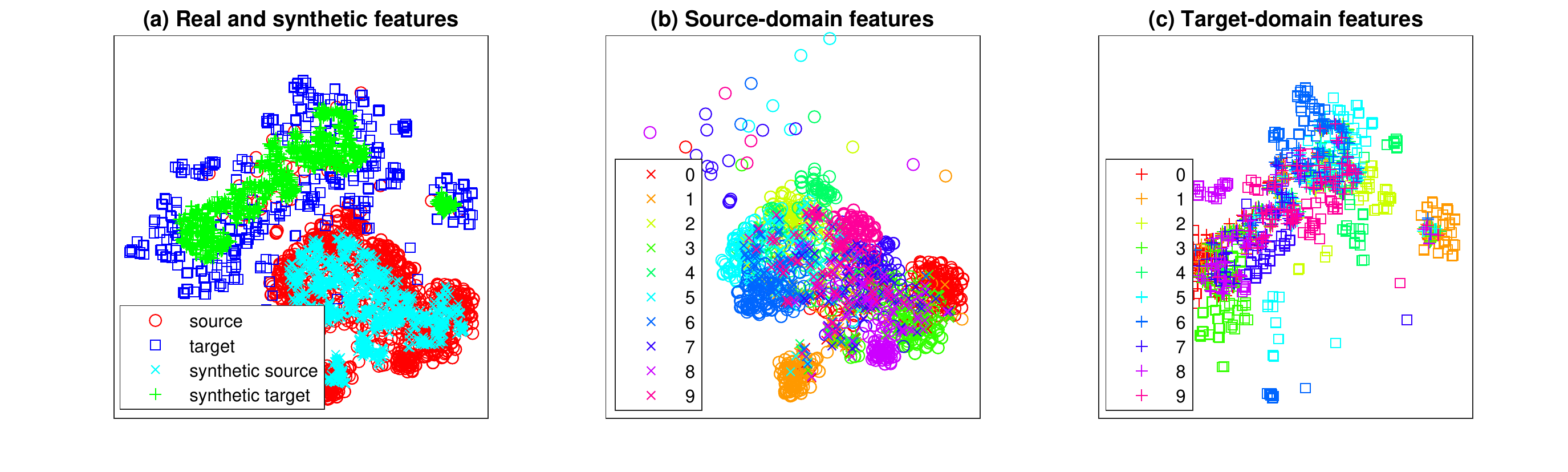}}
	{\caption{Visualization of real and synthetic features using t-SNE (best viewed in color). (a) data distribution of four domains (i.e. real source, real target, synthetic source, synthetic target); (b) real and synthetic data distribution in the source domain (colours represent different classes); (c) real and synthetic data distribution in the target domain (colours represent different classes).}
		\label{fig:visualization}}
\end{figure*}

For qualitative evaluation, we use the t-SNE technique \cite{maaten2008visualizing} to visualize the real and synthetic features in Figure \ref{fig:visualization}. The domain adaptation task $C\to W$ in the Office-Caltech dataset is taken as an exemplar. The 4096-dimensional features of real and synthetic data from both domains are mapped into 2-dimensional projections in an unsupervised way by preserving data distributions \cite{maaten2008visualizing}.

Firstly, we visualize real data points from the source (\textit{red circles}) and target (\textit{blue squares}) domains in Figure \ref{fig:visualization}(a). It is clear data from source and target domains are distributed in different regions. In the same plot, we also visualize the synthetic features generated by our proposed model for the source (\textit{cyan crosses}) and target (\textit{green +}) domains. We can see the synthetic data points generated for the source/target domain are well aligned with the real data points in the corresponding domain thanks to the domain conditions of the decoder in our norm-VAE model.

Secondly, we examine the class discriminative property of synthetic data in the source and target domains in Figure \ref{fig:visualization}(b) and (c) respectively. In the the source domain, we use \textit{circles} and \textit{crosses} to represent the real and synthetic data points respectively whilst different colors are used for ten classes. Similarly, \textit{squares} and \textit{crosses} are used for real and synthetic data points and colors represent different classes in the target domain. We can see that real data points from the same class are distributed in a cluster thanks to the discriminative features extracted by deep CNN models pre-trained on ImageNet. The synthetic data generated by our proposed model are also distributed in clusters of different classes. This demonstrates our proposed method is able to generate synthetic data which are both domain and class discriminative.

Finally, a closer inspection of Figure \ref{fig:visualization}(b) and (c) also tells us that the synthetic data clusters are not perfectly aligned with their corresponding clusters of real data (i.e. circles/squares and crosses of the same colour are not well aligned). Such misalignment is more severe in the target domain due to the fact there is no labelled data in this domain. We believe slight misalignment leads to over-complete data distribution \cite{keshari2020generalized} and is beneficial to learning a more robust classifier. However, significant distribution shifts can have negative effect on the performance. This demonstrates the limitation of our proposed method in generating reliable class-discriminative synthetic data and leads us to improve the model in our future work.  

\section{Conclusion}\label{sec:conclusion}
In this paper, we proposed novel approaches to the unsupervised domain adaptation problem from a novel perspective and achieved impressive experimental results with the average classification accuracy of 97.2\%, 93.4\%, 87.6\%, 90.4\% and 67.9\% on Digits, Office-Caltech, Office31, ImageCLEF-DA and Office-Home datasets, respectively. Instead of pursuing explicit domain adaptation, we train a unified classifier for both source and target domain data in a high-dimensional feature space despite the existence of distribution discrepancy across domains. We proposed a novel pseudo-label selection strategy outperforming the existing ones in the literature \cite{chen2019progressive,wang2019unifying,wang2020unsupervised}. With this specially designed pseudo-labelling strategy, our method {\it naive-SPL} can achieve strong performance which is impressive given that it only uses a typical shallow CNN for digit classification and a linear two-layer MLP for image classification. Moreover, our proposed {\it norm-VAE-SPL} can improve the performance by generating synthetic features for training data augmentation. To conclude, our work provides a fresh insight into unsupervised domain adaptation for the community.

\bibliographystyle{IEEEtran}
\bibliography{ref}

\begin{IEEEbiography}
[{\includegraphics[width=1.2in,height=1.5in,clip,keepaspectratio]{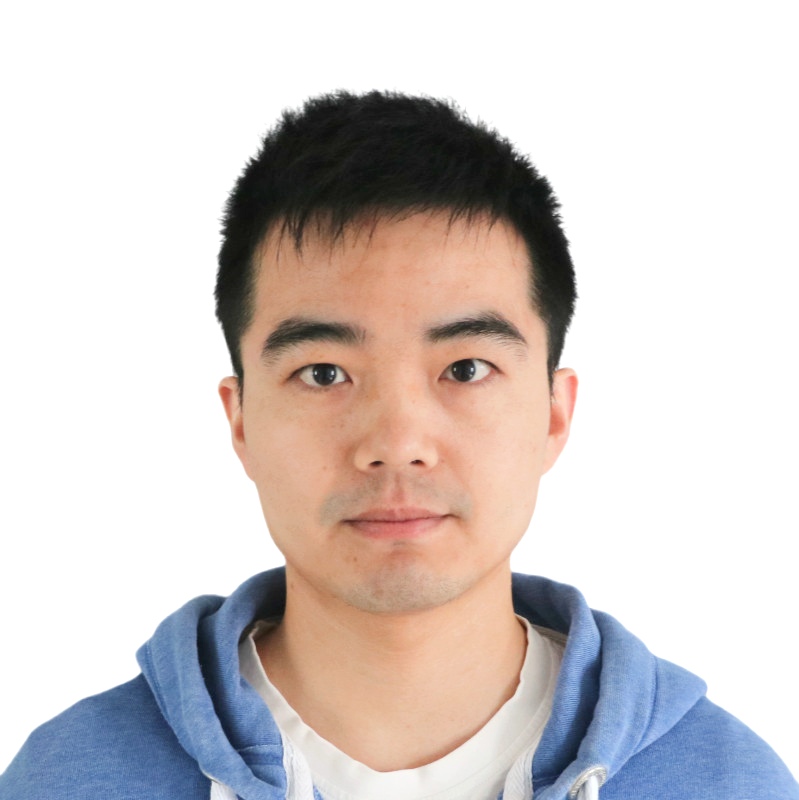}}]{Qian Wang}
is a Research Associate within the Department of Computer Science at Durham University, United Kingdom. His researches focus on deep learning and computer vision. 

Dr. Wang received his PhD in machine learning from The University of Manchester (UK) in 2017, Master's degree in Biomedical Engineering and Bsc in Electronic Engineering in 2013 and 2010 respectively, both from University of Science and Technology of China (Hefei).
\end{IEEEbiography}

\begin{IEEEbiography}
[{\includegraphics[width=1.1in,height=1.5in,clip,keepaspectratio]{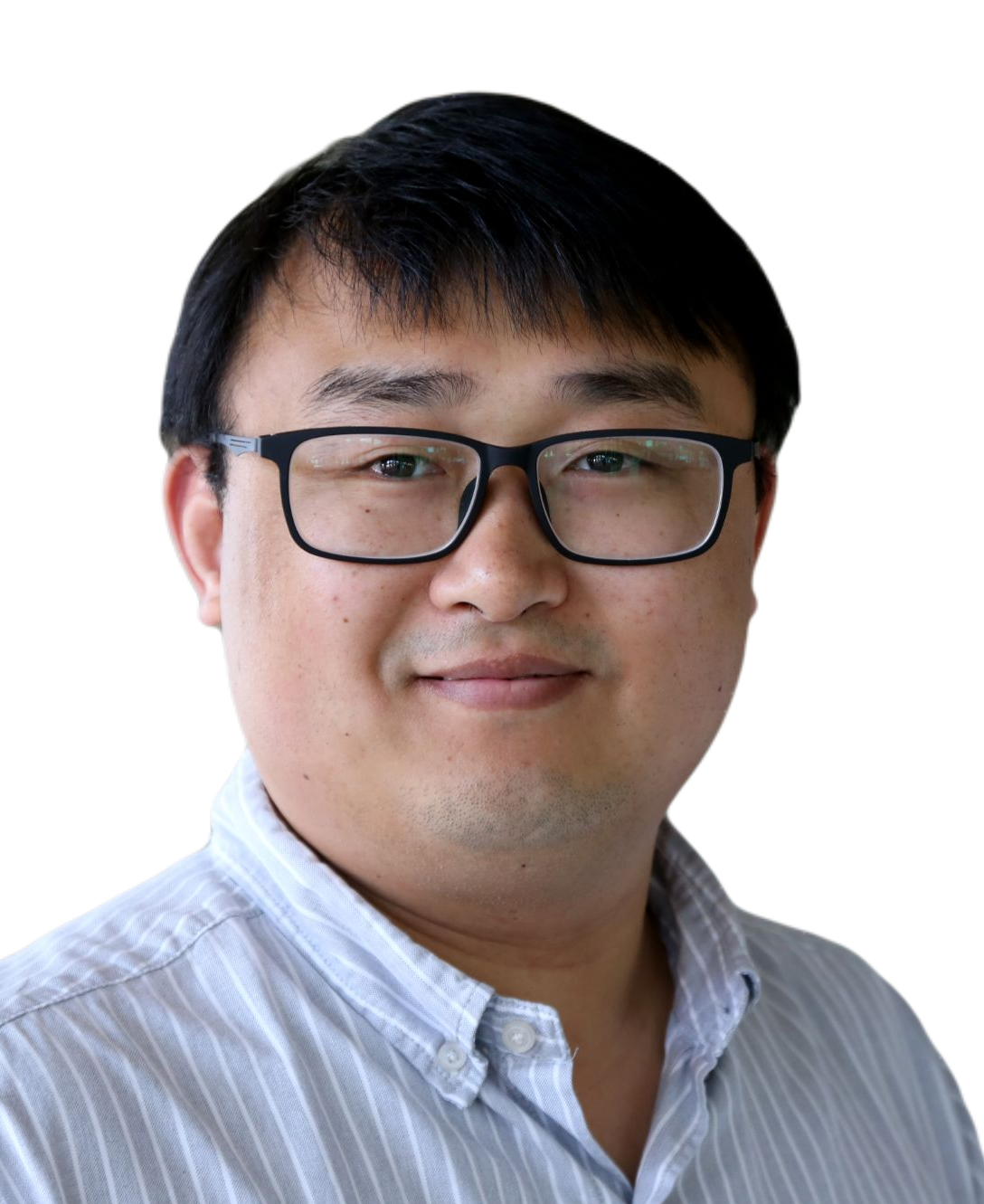}}]{Fanlin Meng}
 is a Lecturer within the Department of Mathematical Sciences at the University of Essex, United Kingdom. His primary research interests are Machine Learning, Game theory and Optimisation including their applications to smart grids, energy markets and intelligent transportation systems. 
 
Dr. Meng received his PhD in Computer Science from the University of Manchester, UK in 2015, MSc in Systems Engineering from Xiamen University, China in 2011 and BSc in Automation from China University of Mining and Technology, China in 2008.  He served in the organising committee of LSMS2020 and ICSEE2020 and the lead organiser of WCCI 2020 and IJCNN 2021 special sessions. 
\end{IEEEbiography}

\begin{IEEEbiography}
[{\includegraphics[width=1.1in,height=1.5in,clip,keepaspectratio]{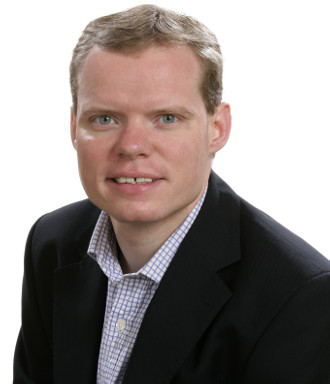}}]{Toby P. Breckon}
is currently a Professor within the Departments of Engineering and Computer Science, Durham University (UK).  His key research interests lie in the domain of computer vision and image processing and he leads a range of research activity in this area.

Prof. Breckon holds a PhD in informatics (computer vision) from the University of Edinburgh (UK).  He has been a visiting member of faculty at the Ecole Supérieure des Technologies Industrielles Avancées (France), Northwestern Polytechnical University (China), Shanghai Jiao Tong University (China) and Waseda University (Japan).

Prof. Breckon is a Chartered Engineer, Chartered Scientist and a Fellow of the British Computer Society. In addition, he is an Accredited Senior Imaging Scientist and Fellow of the Royal Photographic Society. He led the development of image-based automatic threat detection for the 2008 UK MoD Grand Challenge winners [R.J. Mitchell Trophy, (2008), IET Innovation Award (2009)]. His work is recognised as recipient of the Royal Photographic Society Selwyn Award for early-career contribution to imaging science (2011). http://www.durham.ac.uk/toby.breckon/
\end{IEEEbiography}






\end{document}